\documentclass{amia}
\usepackage{graphicx}
\usepackage[labelfont=bf]{caption}
\usepackage[superscript,nomove]{cite}
\usepackage{color}
\usepackage{url}
\usepackage{amsmath}
\usepackage{amsfonts}
\usepackage{amssymb}
\usepackage{booktabs}
\usepackage{wrapfig}
\usepackage{ulem}
\usepackage{tabularx}
\usepackage{enumitem}
\usepackage{comment}
\usepackage{colortbl}
\usepackage{gensymb}
	
\definecolor{Gray}{gray}{0.9}
\definecolor{LightCyan}{rgb}{0.88,1,1}

\begin{document}

\title{Evaluate underdiagnosis and overdiagnosis bias of deep learning model on primary open-angle glaucoma diagnosis in under-served populations}

\author{
    Mingquan Lin$^{1,\text{a}}$,
    Yunyun Xiao$^{1,}$\footnote{Equal contributions.}, 
    Bojian Hou$^{2}$,
    Tingyi Wanyan$^{1}$,
    Mohit Manoj Sharma$^{1}$,\\
    Zhangyang Wang$^{3}$,
    Fei Wang$^{1}$,
    Sarah Van Tassel$^{1}$,
    Yifan Peng$^{1}$
}

\institutes{$^{1}$Population Health Sciences, Weill Cornell Medicine, New York, NY, USA; 
$^{2}$ Department of Biostatistics, Epidemiology and Informatics, University of Pennsylvania;
$^{3}$ Electrical and Computer Engineering, The University of Texas at Austin;
}

\maketitle

\noindent{\bf Abstract}

\textit{%
In the United States, primary open-angle glaucoma (POAG) is the leading cause of blindness, especially among African American and Hispanic individuals. Deep learning has been widely used to detect POAG using fundus images as its performance is comparable to or even surpasses diagnosis by clinicians. However, human bias in clinical diagnosis may be reflected and amplified in the widely-used deep learning models, thus impacting their performance. Biases may cause (1) underdiagnosis, increasing the risks of delayed or inadequate treatment, and (2) overdiagnosis, which may increase individuals’ stress, fear, well-being, and unnecessary/costly treatment. In this study, we examined the underdiagnosis and overdiagnosis when applying deep learning in POAG detection based on the Ocular Hypertension Treatment Study (OHTS) from 22 centers across 16 states in the United States.  Our results show that the widely-used deep learning model can underdiagnose or overdiagnose under-served populations. 
The most underdiagnosed group is female younger ($< 60$ yrs) group, and the most overdiagnosed group is Black older ($\ge 60$ yrs) group. Biased diagnosis through traditional deep learning methods may delay disease detection, treatment and create burdens among under-served populations, thereby, raising ethical concerns about using deep learning models in ophthalmology clinics. 
}

\section{Introduction}

Primary open-angle glaucoma (POAG) is one of the leading causes of blindness in the US and worldwide.\cite{Bourne2013-dn}
It has been projected to affect approximately 111.8 million people by 2040. Among these patients, 5.3 million may be bilaterally blind.\cite{Quigley2006-hc}
In the United States, POAG is the most common form of glaucoma and is the leading cause of blindness among Black and Hispanic individuals.\cite{Jiang2018-rq, Sommer1991-rc}
POAG is asymptomatic until it reaches an advanced stage whereby peripheral visual field loss encroaches on central vision. However, early detection and treatment can avoid most blindness caused by POAG.\cite{Tatham2015-ef}
Therefore, timely identification of individuals with glaucoma is critical to guide  medical and surgical treatments and patient monitoring.\cite{Doshi2008-iy, Quigley1992-xi}
Detecting POAG is crucial, yet challenging, due to the high demands on screening and experiences of ophthalmologists.\cite{Kolomeyer2021-pa}
Fundus photography is convenient and inexpensive for recording optic nerve head structure \cite{quigley1986examination}. Therefore, developing an automatic deep learning (DL) model to detect POAG with high accuracy from fundus photographs to address the lack of experienced ophthalmologists is important.

In recent years, developments in artificial intelligence (AI) have provided potential opportunities for automatic POAG diagnosis using fundus photographs.\cite{Raghavendra2018-ej, Li2020-wd, Li2018-uv, Thakur2020-ip, Fu2018-ja, Fan2022-ww, lin2022primary, lin2022automated}  While such DL models have increasingly achieved expert-level performance, there is growing concern that these models may reflect and amplify human bias and reduce the quality of their per- formance in historically under-served populations such as Black individuals.\cite{Obermeyer2019-ho, Chen2020-om, Char2018-fu, Wiens2019-km} The topic of AI-driven underdiagnosis and overdiagnosis has been particularly important.\cite{seyyed2020chexclusion, Wiens2019-km} In this work, we define “\emph{Underdiagnosis}” as the problem where the model falsely claims that the individual is healthy, increasing the risks of delayed or inadequate treatment, and “\emph{Overdiagnosis}” as the problem where the model predicts healthy people wrongly into sick ones, leading to individuals’ stress, fear, and unnecessary/costly treatment. Therefore, model fairness of POAG diagnosis can be a crucial concern if used in the clinical pipeline for patient triage.

To our knowledge, the topic of AI-driven underdiagnosis and overdiagnosis of POAG diagnosis has not been explored before.\cite{XU2022104250} In this study, we will systematically study underdiagnosis and overdiagnosis bias in the AI-based POAG diagnosis models from a large, cross-sectional dataset obtained from the Ocular Hypertension Treatment Study (OHTS).\cite{Kass2002-jv}  OHTS is one of the largest longitudinal clinical trials in POAG (1,636 participants and 37,399 images) from 22 centers in the United States. \cite{Gordon1999-om} 
We chose these subgroups mainly because of the clear history of bias in previous research.\cite{Larrazabal2020-sx, Vyas2020-pt, Sun2020-no, Seyyed-Kalantari2021-tt}

\section{Materials and Methods}

\subsection{The OHTS dataset}

In this study, we perform a study of model bias in POAG diagnosis in a large-scale, longitudinal, and population-based dataset (Table~\ref{table:data}). Previous studies show that the classifier exhibits different performances in different individual groups stratified by sex, race, and age.\cite{Seyyed-Kalantari2021-tt} Inspired by these works, we report results by considering these factors. 

The dataset is obtained from the Ocular Hypertension Treatment Study (OHTS). The study protocol was approved by the Institutional Review Board at each clinical center, and the Weill Cornell Medicine IRB determined that the protocol does not constitute human subjects research. 
All risk factors were measured at baseline before the onset of the disease and collected for approximately 16 years. The participants in this dataset were selected according to both eligibility and exclusion criteria.\cite{Kass2002-jv}

Briefly, the eligibility criteria include intraocular pressure (between 24 mm Hg and 32 mm Hg in one eye and between 21 mm Hg and 32 mm Hg in the fellow eye) and age (between 40 and 80 years old). The visual field tests were interpreted by the Visual Field Reading Center, and the stereoscopic photographs were interpreted by the Optic Disc Reading Center. Exclusion criteria included previous intraocular surgery, visual acuity worse than 20/40 in either eye, and diseases that may cause optic disc deterioration and visual field loss (such as diabetic retinopathy). The gold standard POAG labels were graded at the Optic Disc Reading Center. In brief, two masked certified readers were instructed to independently detect glaucomatous optic disc deterioration over time. If there was a disagreement between two readers, a senior reader reviewed the subject in a masked fashion. The POAG diagnosis in a quality control sample of 86 eyes (50 normal eyes and 36 with progression) showed test-retest agreement at $\kappa = 0.70$ (95\% confidence interval [CI], 0.55-0.85). More details of the reading center workflow have been described in Gorden et al.\cite{Gordon1999-om}

\begin{table}[htbp]
\centering
\begin{tabular}{lr@{\hspace{4em}}r@{\hspace{4em}}r@{~~(}r@{)}}
\toprule
\hspace{2em} & & Total & \multicolumn{2}{r}{POAG}\\
 \midrule
\multicolumn{2}{l}{No. of images}  & 37,399 & 2,327 & 6.22\%\\
\rowcolor{Gray}
\multicolumn{5}{l}{Sex}\\
& Male  & 16,185 & 1,303  & 8.05\%\\
& Female & 21,154 & 1,024 & 4.84\% \\
\rowcolor{Gray}
\multicolumn{5}{l}{Race}\\
& Non-Black & 28,460 & 1,554 & 5.46\% \\
& Black & 8,879 & 773 & 8.71\% \\
\rowcolor{Gray}
\multicolumn{5}{l}{Age} \\
& 40-49 & 4,292 & 64 & 1.49\%\\
& 50-59	& 11,962 & 356 & 2.98\%\\
& 60-69	& 11,904 & 846 & 7.11\%\\
& 70-79	& 7,593 & 829 & 10.92\%\\
& $\ge$80	& 1,588 & 232 & 14.61\%\\
\bottomrule
\end{tabular}
\caption{The characteristics of the OHTS dataset.}
\label{table:data}
\end{table}

\subsection{Definition of POAG underdiagnosis and overdiagnosis}

To assess model fairness, we compare underdiagnosis and overdiagnosis rates across subpopulations. Similar to ref,\cite{Seyyed-Kalantari2021-tt} we define the underdiagnosis rate as the false-negative rate (FNR) of the binarized model prediction for the POAG at the subgroup levels: $P(\hat{y}=\text{non-POAG} | y=\text{POAG},A)$. Here $A$ is the sex, race, or other factors that the model should be free of bias. For example, the underdiagnosis of female individuals is given by $P(\hat{y}=\text{non-POAG} | y=\text{POAG},\text{female})$. We then compare these underdiagnosis rates across subpopulations. We say a classifier is fair if the individuals in the protected and unprotected groups satisfy the formula: 
\begin{equation*}
 P(\hat{y}=\text{non-POAG} | y=\text{POAG},\text{female})=P(\hat{y}=\text{non-POAG} | y=\text{POAG},\text{male})   
\end{equation*}

For overdiagnosis, we will measure the false-positive rate (FPR) for the POAG across all subgroups, i.e., $P(\hat{y}=\text{POAG} | y=\text{non-POAG},A)$. This measure shows that the model fails to diagnose those individuals who would never develop POAG. Some of the harms caused by overdiagnosis are anxiety and having treatments that are not needed.

Besides single identities, we also examined underdiagnosis and overdiagnosis in \emph{intersectional} groups - individuals who belong to two subpopulations.\cite{Seyyed-Kalantari2021-tt} For example, the underdiagnosis of Black female individuals is given by  $P(\hat{y}=\text{non-POAG} | y=\text{POAG},\text{female},\text{Black})$. Here, we want to examine if individuals who belong to two subgroups may have a larger underdiagnosis rate. In other words, not all female individuals are misdiagnosed at the same rate (for example, Black female individuals are misdiagnosed more than non-Black female individuals).

\subsection{Model development}

\begin{figure}
\centering
\includegraphics[width=.7\textwidth]{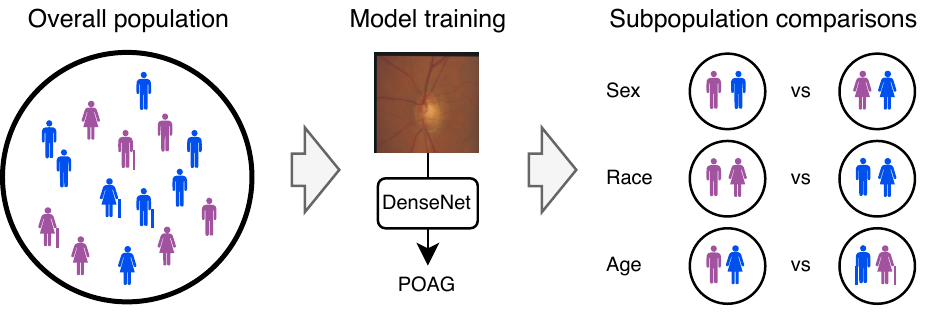}
\caption{The experimental design. We focus our underdiagnosis and overdiagnosis experiments on subgroups of race, sex, and age. 
}
\label{pipeline}
\end{figure}

Figure~\ref{pipeline} shows the pipeline of our model.
All images are resized to  $224 \times 224 \times 3$ and normalized using the mean and standard deviation of the ImageNet dataset.\cite{Deng2009-hb} 
We sequentially apply three augmentation operations on the fly during training: (1) random rotation between $0\degree$ and $10\degree$, (2) random translation: an image was translated randomly along the x- and y-axes by distances ranging from 0 to 10\% of width or height of the image, and (3) random flipping. The diversity of the dataset could be increased due to these data augmentation techniques, which generate effective and robust representations.

The input images are then passed through a convolutional neural network to generate the prediction results. In this study, we used the DenseNet-201\cite{huang2017densely} pre-trained on ImageNet.\cite{russakovsky2015imagenet}
We replaced the last layer with a new randomly initialized fully-connected layer with 2 output neurons (POAG and normal). We used binary cross-entropy as the loss function. 
Since there are only 6.22\% of images in the OHTS dataset that has POAG (Table~\ref{table:data}, a severe class imbalance exists for POAG diagnosis, to overcome this problem, we adopted weighted cross-entropy, a commonly used loss function in classification. The adopted weighted cross-entropy was: 
\begin{equation}
    \mathcal{L_{\theta}}=-\frac{1}{N}\sum_{i=1}^N[\beta y_i\log(\hat{y}_i \text{DenseNet}(x_i,\theta))+(1-\beta)(1-y_i)\log(1-\hat{y}_i \text{DenseNet}(x_i,\theta))]
\end{equation}
where $N$ is the number of training examples, $\beta$ is the balancing factor between positive and negative samples, $y_i$ is the observed true label of image $x_i$, $\hat{y}_i$ is the probability predicted by the classifier, and $\theta_s$ represents the parameters of the DenseNet-201. Here, we used inversely proportional to POAG frequency in the training data. Finally, we fine-tuned the entire network on the OHTS in an end-to-end manner.

\subsection{Experimental settings}

The model was implemented by Keras with a backend of Tensorflow. The network was optimized using the Adam optimizer method.\cite{kingma2014adam} The learning rate is $5 \times 10^{-5}$. The experiments were performed on Intel Core i9-9960 X 16 cores processor and NVIDIA Quadro RTX 6000 GPU. 

We used the five-fold cross-validation in this study. We split the entire dataset randomly into five groups at the individual level. This ensured that no participant was in more than one group to avoid cross-contamination between the training and testing datasets. In each fold of the cross-validation, we took one group (20\% of total subjects) as the hold-out test set and the remaining 4 groups as the training set.

\section{Results and discussion}
\label{sec:results}

The underdiagnosis (Figure \ref{result}) and overdiagnosis (Figure \ref{result2}) for POAG screening show an inverse relationship in both subgroups and intersectional groups in the OHTS dataset. As suggested by Seyyed et al,\cite{Seyyed-Kalantari2021-tt} this indicates that the model consistently misclassifies the under-served subpopulations due to potential biases, rather than simple, random errors.

\subsection{Underdiagnosis in subpopulations and intersectional groups}

\begin{figure}
\centering
\includegraphics[width=.7\textwidth]{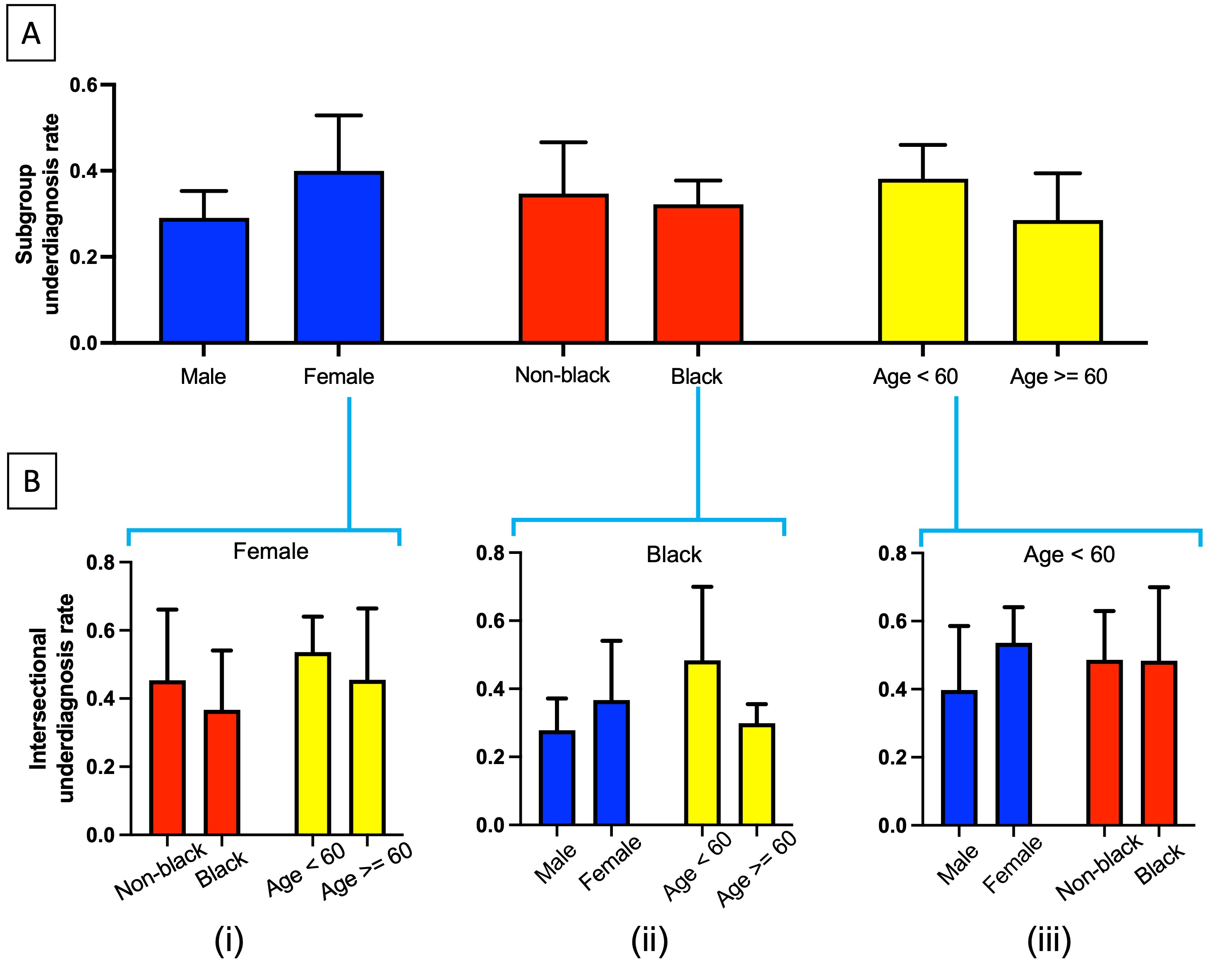}
\caption{Underdiagnosis analysis across subgroups of sex, race, and age in the OHTS dataset. A) The underdiagnosis rates in the subpopulations. B) Intersectional underdiagnosis rates for female individuals, Black individuals, and individuals aged $<$ 60 years, respectively. The results are average results of the five-fold cross-validation.}\label{result}
\end{figure}

Figure \ref{result}A shows that the underdiagnosis rate differs in all subpopulations of sex, race, and age. Specifically, females, non-Black individuals, and individuals under 60 years old have higher underdiagnosis rates than their counterparts. In other words, the individuals of these groups are more likely falsely predicted as healthy, preventing them from receiving appropriate treatments. 
From Table~\ref{table:data}, we can see that the individuals in these groups have a lower prevalence of POAG. For example, the POAG rate of individuals aged $\ge 60$ is around three times more than that of individuals aged $<60$ (9.04\% vs. 2.58\%). Since these groups may not be adequately represented in the OHTS data, the supervised machine learning model trained from the data might be biased.  

In addition, female individuals have the highest underdiagnosis rate among indicated subgroups. However, the numbers of POAG female and male individuals are about the same. This observation suggests that a simple resampling approach to ensure the dataset is balanced across different groups may not always be a solution. 

We also investigate intersectional groups, which means the individuals belong to two subpopulations, e.g., female Black individuals. As shown in Figure~\ref{result}B(i)-(iii), intersectional groups also have the problem of underdiagnosis. 
Figure \ref{result}B(i) shows that female non-Black individuals have a higher underdiagnosis rate than female Black individuals. Female individuals aged $<60$ years have a higher rate than female individuals aged $\ge 60$ years. Compared to the single identities,  we observed that the intersectional identities amplify the model bias. For example, the difference between Black and non-Black individuals is 2.50\%, but the difference between female Black and female non-Black is 8.61\%. The most underdiagnosed group is young female individuals.

Similarly, Figure~\ref{result}B(ii) shows that Black females and younger individuals have a higher underdiagnosis rate than Black males and older adults. Figure~\ref{result}B(iii) also indicates that female youngers are more heavily underdiagnosed. There is no significant difference between Black and non-Black younger individuals. This may be partially due to the small test set sizes (84 cases with the POAG label for individuals aged $<60$ years).

\subsection{Overdiagnosis in subpopulations and intersectional groups}

\begin{figure}
\centering
\includegraphics[width=.7\textwidth]{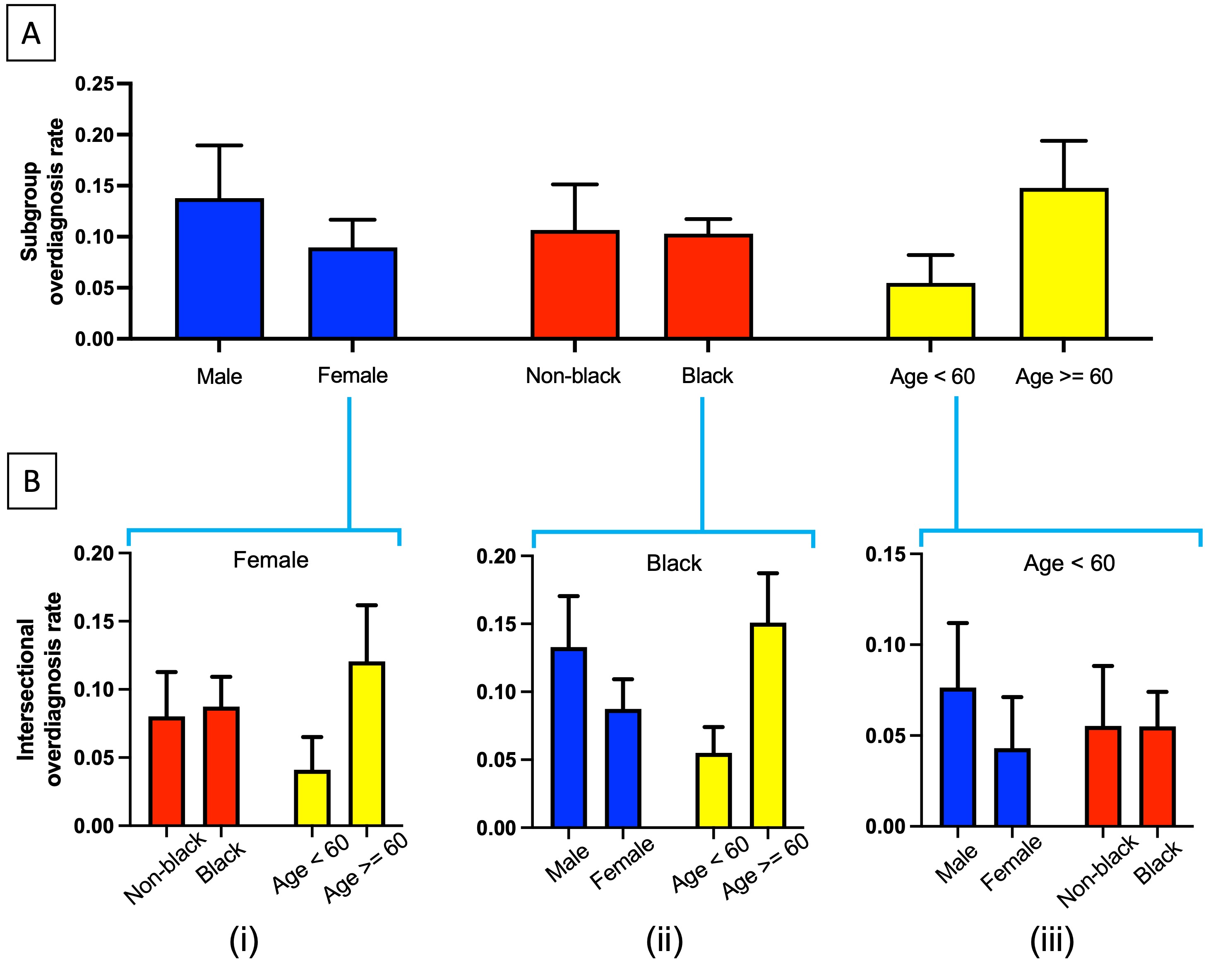}
\caption{Overdiagnosis analysis across subgroups of sex, race, and age in the OHTS dataset. A) The overdiagnosis rates in the subpopulations. B) Intersectional overdiagnosis rates for female individuals, Black individuals, and individuals aged $<$ 60 years, respectively. The results are average results of the five-fold cross-validation.}\label{result2}
\end{figure}

Figure \ref{result2}A shows that healthy males and healthy older adults are more likely to be misclassified as POAG positive.
On the other hand, the Black and non-Black subpopulations in Figure \ref{result2}A have similar overdiagnosis rates.
%
From Figure \ref{result2}B, we observed that the intersectional identities are often overdiagnosed  more heavily than the group in aggregate. Specifically, female and Black older adults are more easily overdiagnosed than female and Black younger adults (Figure \ref{result2}B(i) and (ii)).

\subsection{Discussions and Limitations}

We have found that underdiagnosis and overdiagnosis exist in the OHTS dataset for POAG diagnosis. The DL model generated underdiagnosis and overdiagnosis biases in under-served subpopulations, such as female individuals and individuals under 60 years old. Such biased diagnoses were even greater among individuals with intersectional identities, including Black females and females aged under 60 years old. Subpopulations like female individuals and individuals under 60 years old are most affected by POAG in the OHTS dataset, suggesting further attention when applying DL models in clinical decision-making\cite{hou2021clinical}.

The DL model performs similarly on Black and non-Black individuals mainly because the number of Black individuals in our datasets is much smaller than the other group, even though they have a higher prevalence of POAG. Therefore, the adverse effects on Black individuals are probably mitigated by the small number of their population. This reminds us that we must also consider the number of subpopulations when constructing the dataset\cite{mehrabi2021survey}. 

We also found that these differences in underdiagnosis and overdiagnosis exist in other clinical research areas (e.g., thoracic diseases, heart diseases, and kidney diseases),\cite{Seyyed-Kalantari2021-tt,mamary2018race,sun2020exploring} which means that these disparities may be widespread in biomedical study.

One limitation of this study is that only one dataset was used. However, the OHTS data is one of the largest longitudinal clinical trials in POAG from 22 centers in the United States. Therefore, we believe the observations of model bias are likely to be generalizable. 

Another limitation is that we only studied the fairness of a binarized model. Unfortunately, the probabilities predicted by the model may not be calibrated: probabilities are calibrated where a prediction of POAG with confidence $p$ is correct $p$ percent of the time.\cite{pleiss2017fairness}
That being said, the probabilities predicted by the model may be over-confident in some cases and under-confident in other cases.  Moreover, as we see in Figures~\ref{result}, the severely imbalanced data may result in even more bias in the predicted probabilities as they over-favor in predicting the majority class. As such, we should investigate the relationship between calibration and POAG underdiagnosis/overdiagnosis in the future. In addition, we plan to develop an efficient method to reduce bias in the future.

\section{Conclusion}

In this paper, we systematically study underdiagnosis and overdiagnosis bias in the DL-based POAG diagnosis models and identify the factors contributing to model fairness. We find deep learning-based underdiagnosis and overdiagnosis exist among under-served subpopulations in POAG diagnosis on the OHTS dataset.  Underdiagnosis will prevent the individuals from receiving appropriate treatment, while overdiagnosis will let the individuals receive or continue receiving unnecessary treatments. Bias between the individuals in intersectional subgroups such as females under 60 years and Black females are more severe. This emphasizes that bias mitigation approaches should consider the combination of characteristics involved in bias rather than a single identity. As deep learning models are implemented in clinical practice, this problem takes on particular urgency.



\section*{Acknowledgment}

This work was supported by the National Library of Medicine under Award No. 4R00LM013001, NSF CAREER Award No. 2145640, and Amazon Research Award.

\makeatletter
\renewcommand{\@biblabel}[1]{\hfill #1.}
\makeatother

\bibliography{bibm_bib,bib2}

\begin{thebibliography}{10}

\bibitem{Bourne2013-dn}
Bourne RRA, Stevens GA, White RA, Smith JL, Flaxman SR, Price H, et~al.
\newblock Causes of vision loss worldwide, 1990-2010: a systematic analysis.
\newblock Lancet Glob Health. 2013 Dec;1(6):e339-49.

\bibitem{Quigley2006-hc}
Quigley HA, Broman AT.
\newblock The number of people with glaucoma worldwide in 2010 and 2020.
\newblock Br J Ophthalmol. 2006 Mar;90(3):262-7.

\bibitem{Jiang2018-rq}
Jiang X, Torres M, Varma R, {Los Angeles Latino Eye Study Group}.
\newblock Variation in Intraocular Pressure and the Risk of Developing
  {Open-Angle} Glaucoma: The Los Angeles Latino Eye Study.
\newblock Am J Ophthalmol. 2018 Apr;188:51-9.

\bibitem{Sommer1991-rc}
Sommer A, Tielsch JM, Katz J, Quigley HA, Gottsch JD, Javitt JC, et~al.
\newblock Racial differences in the cause-specific prevalence of blindness in
  east Baltimore.
\newblock N Engl J Med. 1991 Nov;325(20):1412-7.

\bibitem{Tatham2015-ef}
Tatham AJ, Medeiros FA, Zangwill LM, Weinreb RN.
\newblock Strategies to improve early diagnosis in glaucoma.
\newblock Prog Brain Res. 2015 Jul;221:103-33.

\bibitem{Doshi2008-iy}
Doshi V, Ying-Lai M, Azen SP, Varma R, {Los Angeles Latino Eye Study Group}.
\newblock Sociodemographic, family history, and lifestyle risk factors for
  open-angle glaucoma and ocular hypertension. The Los Angeles Latino Eye
  Study.
\newblock Ophthalmology. 2008 Apr;115(4):639-47.e2.

\bibitem{Quigley1992-xi}
Quigley HA, Katz J, Derick RJ, Gilbert D, Sommer A.
\newblock An evaluation of optic disc and nerve fiber layer examinations in
  monitoring progression of early glaucoma damage.
\newblock Ophthalmology. 1992 Jan;99(1):19-28.

\bibitem{Kolomeyer2021-pa}
Kolomeyer NN, Katz LJ, Hark LA, Wahl M, Gajwani P, Aziz K, et~al.
\newblock Lessons Learned From 2 Large Community-based Glaucoma Screening
  Studies.
\newblock J Glaucoma. 2021 Oct;30(10):875-7.

\bibitem{quigley1986examination}
Quigley HA.
\newblock Examination of the retinal nerve fiber layer in the recognition of
  early glaucoma damage.
\newblock Transactions of the American Ophthalmological Society. 1986;84:920.

\bibitem{Raghavendra2018-ej}
Raghavendra U, Fujita H, Bhandary SV, Gudigar A, Tan JH, Acharya UR.
\newblock Deep convolution neural network for accurate diagnosis of glaucoma
  using digital fundus images.
\newblock Inf Sci. 2018 May;441:41-9.

\bibitem{Li2020-wd}
Li L, Xu M, Liu H, Li Y, Wang X, Jiang L, et~al.
\newblock A {Large-Scale} Database and a {CNN} Model for {Attention-Based}
  Glaucoma Detection.
\newblock IEEE Trans Med Imaging. 2020 Feb;39(2):413-24.

\bibitem{Li2018-uv}
Li Z, He Y, Keel S, Meng W, Chang RT, He M.
\newblock Efficacy of a Deep Learning System for Detecting Glaucomatous Optic
  Neuropathy Based on Color Fundus Photographs.
\newblock Ophthalmology. 2018 Aug;125(8):1199-206.

\bibitem{Thakur2020-ip}
Thakur A, Goldbaum M, Yousefi S.
\newblock Predicting Glaucoma before Onset Using Deep Learning.
\newblock Ophthalmol Glaucoma. 2020 Jul;3(4):262-8.

\bibitem{Fu2018-ja}
Fu H, Cheng J, Xu Y, Zhang C, Wong DWK, Liu J, et~al.
\newblock {Disc-Aware} Ensemble Network for Glaucoma Screening From Fundus
  Image.
\newblock IEEE Trans Med Imaging. 2018 Nov;37(11):2493-501.

\bibitem{Fan2022-ww}
Fan R, Bowd C, Christopher M, Brye N, Proudfoot JA, Rezapour J, et~al.
\newblock Detecting Glaucoma in the Ocular Hypertension Study Using Deep
  Learning.
\newblock JAMA Ophthalmol. 2022 Apr;140(4):383-91.

\bibitem{lin2022primary}
Lin M, Liu L, Gordon M, Kass M, Wang F, Van~Tassel SH, et~al.
\newblock Primary open-angle glaucoma diagnosis from optic disc photographs
  using Siamese network.
\newblock Ophthalmology Science. 2022:100209.

\bibitem{lin2022automated}
Lin M, Hou B, Liu L, Gordon M, Kass M, Wang F, et~al.
\newblock Automated diagnosing primary open-angle glaucoma from fundus image by
  simulating human’s grading with deep learning.
\newblock Scientific reports. 2022;12(1):1-10.

\bibitem{Obermeyer2019-ho}
Obermeyer Z, Powers B, Vogeli C, Mullainathan S.
\newblock Dissecting racial bias in an algorithm used to manage the health of
  populations.
\newblock Science. 2019 Oct;366(6464):447-53.

\bibitem{Chen2020-om}
Chen IY, Joshi S, Ghassemi M.
\newblock Treating health disparities with artificial intelligence.
\newblock Nat Med. 2020 Jan;26(1):16-7.

\bibitem{Char2018-fu}
Char DS, Shah NH, Magnus D.
\newblock Implementing machine learning in health care—addressing ethical
  challenges.
\newblock The New England journal of medicine. 2018;378(11):981.

\bibitem{Wiens2019-km}
Wiens J, Saria S, Sendak M, Ghassemi M, Liu VX, Doshi-Velez F, et~al.
\newblock Do no harm: a roadmap for responsible machine learning for health
  care.
\newblock Nat Med. 2019 Sep;25(9):1337-40.

\bibitem{seyyed2020chexclusion}
Seyyed-Kalantari L, Liu G, McDermott M, Chen IY, Ghassemi M.
\newblock CheXclusion: Fairness gaps in deep chest X-ray classifiers.
\newblock In: BIOCOMPUTING 2021: proceedings of the Pacific symposium. World
  Scientific; 2020. p. 232-43.

\bibitem{XU2022104250}
Xu J, Xiao Y, Wang WH, Ning Y, Shenkman EA, Bian J, et~al.
\newblock Algorithmic fairness in computational medicine.
\newblock eBioMedicine. 2022;84:104250.

\bibitem{Kass2002-jv}
Kass MA, Heuer DK, Higginbotham EJ, Johnson CA, Keltner JL, Miller JP, et~al.
\newblock The Ocular Hypertension Treatment Study: a randomized trial
  determines that topical ocular hypotensive medication delays or prevents the
  onset of primary open-angle glaucoma.
\newblock Arch Ophthalmol. 2002 Jun;120(6):701-13; discussion 829-30.

\bibitem{Gordon1999-om}
Gordon MO, Kass MA.
\newblock The Ocular Hypertension Treatment Study: design and baseline
  description of the participants.
\newblock Arch Ophthalmol. 1999 May;117(5):573-83.

\bibitem{Larrazabal2020-sx}
Larrazabal AJ, Nieto N, Peterson V, Milone DH, Ferrante E.
\newblock Gender imbalance in medical imaging datasets produces biased
  classifiers for computer-aided diagnosis.
\newblock Proc Natl Acad Sci U S A. 2020 Jun;117(23):12592-4.

\bibitem{Vyas2020-pt}
Vyas DA, Eisenstein LG, Jones DS.
\newblock Hidden in plain sight—reconsidering the use of race correction in
  clinical algorithms.
\newblock New England Journal of Medicine. 2020;383(9):874-82.

\bibitem{Sun2020-no}
Sun TY, Walk~IV OJ, Chen JL, Nieva HR, Elhadad N.
\newblock Exploring gender disparities in time to diagnosis.
\newblock arXiv preprint arXiv:201106100. 2020.

\bibitem{Seyyed-Kalantari2021-tt}
Seyyed-Kalantari L, Zhang H, McDermott MBA, Chen IY, Ghassemi M.
\newblock Underdiagnosis bias of artificial intelligence algorithms applied to
  chest radiographs in under-served patient populations.
\newblock Nat Med. 2021 Dec;27(12):2176-82.

\bibitem{Deng2009-hb}
Deng J, Dong W, Socher R, Li LJ, Li K, Fei-Fei L.
\newblock {ImageNet}: A large-scale hierarchical image database.
\newblock In: 2009 {IEEE} Conference on Computer Vision and Pattern
  Recognition; 2009. p. 248-55.

\bibitem{huang2017densely}
Huang G, Liu Z, Van Der~Maaten L, Weinberger KQ.
\newblock Densely connected convolutional networks.
\newblock In: Proceedings of the IEEE conference on computer vision and pattern
  recognition; 2017. p. 4700-8.

\bibitem{russakovsky2015imagenet}
Russakovsky O, Deng J, Su H, Krause J, Satheesh S, Ma S, et~al.
\newblock Imagenet large scale visual recognition challenge.
\newblock International journal of computer vision. 2015;115(3):211-52.

\bibitem{kingma2014adam}
Kingma DP, Ba J.
\newblock Adam: A method for stochastic optimization.
\newblock arXiv preprint arXiv:14126980. 2014.

\bibitem{hou2021clinical}
Hou B, Zhang H, Ladizhinsky G, Yang S, Kuleshov V, Wang F, et~al.
\newblock Clinical evidence engine: proof-of-concept for a
  clinical-domain-agnostic decision support infrastructure.
\newblock arXiv preprint arXiv:211100621. 2021.

\bibitem{mehrabi2021survey}
Mehrabi N, Morstatter F, Saxena N, Lerman K, Galstyan A.
\newblock A survey on bias and fairness in machine learning.
\newblock ACM Computing Surveys (CSUR). 2021;54(6):1-35.

\bibitem{mamary2018race}
Mamary AJ, Stewart JI, Kinney GL, Hokanson JE, Shenoy K, Dransfield MT, et~al.
\newblock Race and gender disparities are evident in COPD underdiagnoses across
  all severities of measured airflow obstruction.
\newblock Chronic Obstructive Pulmonary Diseases: Journal of the COPD
  Foundation. 2018;5(3):177.

\bibitem{sun2020exploring}
Sun TY, Walk~IV OJ, Chen JL, Nieva HR, Elhadad N.
\newblock Exploring gender disparities in time to diagnosis.
\newblock arXiv preprint arXiv:201106100. 2020.

\bibitem{pleiss2017fairness}
Pleiss G, Raghavan M, Wu F, Kleinberg J, Weinberger KQ.
\newblock On fairness and calibration.
\newblock Advances in neural information processing systems. 2017;30.

\end{thebibliography}
\bibliographystyle{vancouver}

\end{document}